\documentclass[10pt,twocolumn,letterpaper]{article}

\usepackage[pagenumbers]{cvpr} 

\definecolor{cvprblue}{rgb}{0.21,0.49,0.74}

\newcommand{\myPara}[1]{\vspace{0.04in}\noindent\textbf{#1}\quad}
\definecolor{lightgray}{gray}{0.91}
\usepackage{multirow}
\usepackage{colortbl}
\usepackage{fontawesome} 
\usepackage{amssymb}
\definecolor{upcolor}{RGB}{82,158,63}
\definecolor{downcolor}{RGB}{239,134,54}

\newcommand{\redscore}[2]{$\text{#1}_{\text{\footnotesize{\color{downcolor}{#2}}}}$}
\newcommand{\greenscore}[2]{$\text{#1}_{\textbf{\text{\footnotesize{\color{upcolor}{#2}}}}}$}

\usepackage{algorithm}
\usepackage{algpseudocode}
\usepackage[accsupp]{axessibility}  

\usepackage[pagebackref,breaklinks,colorlinks,allcolors=cvprblue]{hyperref}

\def\paperID{} 
\def\confName{CVPR}
\def\confYear{2026}

\title{Predictive Regularization Against Visual Representation Degradation in Multimodal Large Language Models}

\author{
Enguang Wang$^{1,2, }$\thanks{This work was done during an internship at Tencent Youtu Lab.}\quad
Qiang Wang$^{4}$\quad
Yuanchen Wu$^{4}$\quad
Ke Yan $^{4, \text{\dag}}$\quad 
Xinbin Yuan$^{1,2}$\quad \\
Shouhong Ding$^{4}$\quad
 Xialei Liu$^{1,2,3,}$\thanks{Corresponding author.}\quad 
Ming-Ming Cheng$^{1,2,3}$\quad
\\
\scriptsize{
$^{1}$ NKIARI, Shenzhen Futian \quad
$^{2}$ VCIP, CS, Nankai University \quad
$^{3}$ AAIS, Nankai University \quad
$^{4}$ Tencent Youtu Lab} \\
{\tt\scriptsize enguangwang@mail.nankai.edu.cn\quad \ kerwinyan@tencent.com\quad \ xialei@nankai.edu.cn }
}

\begin{document}
\maketitle
\begin{abstract}

While Multimodal Large Language Models (MLLMs) excel at vision-language tasks, the cost of their language-driven training on internal visual foundational competence remains unclear.   In this paper, we conduct a detailed diagnostic analysis to unveil a pervasive issue: visual representation degradation in MLLMs.   Specifically, we find that compared to the initial visual features, the visual representation in the middle layers of LLM exhibits both a degradation in global function and patch structure.   
We attribute this phenomenon to a visual sacrifice driven by the singular text-generation objective, where the model compromises its visual fidelity to optimize for answer generation.
We argue that a robust MLLM requires both strong cross-modal reasoning and core visual competence, and propose Predictive Regularization (PRe)  to force degraded intermediate features to predict initial visual features, thereby maintaining the inherent visual attributes of the MLLM's internal representations.  Extensive experiments confirm that mitigating this visual degradation effectively boosts vision-language performance, underscoring the critical importance of fostering robust internal visual representations within MLLMs for comprehensive multimodal understanding.

\end{abstract}    
\section{Introduction}
\label{sec:intro}
Recent years have witnessed a monumental leap in artificial intelligence, a paradigm shift largely driven by the unprecedented success of Large Language Models (LLMs)~\cite{devlin2019bert,radford2018improving,radford2019language,touvron2023llama,bai2023qwen,2023internlm,liu2024deepseek}. These models, typically built upon the Transformer architecture~\cite{vaswani2017attention}, are pre-trained on vast text corpora to acquire a remarkable ability to comprehend, reason, and generate human-like text.
Building upon this powerful linguistic foundation, researchers have begun integrating visual information into LLMs, bridging the semantic gap between vision and language to create Multimodal Large Language Models (MLLMs)~\cite{liu2024llavanext, liu2023improvedllava, chen2024internvl,wang2025internvl3_5,lu2024deepseekvl,Qwen-VL,Qwen2.5-VL}. 
By conditioning the language model's generative process on visual inputs, MLLMs are capable of performing a diverse array of vision-language tasks~\cite{Hudson_2019_CVPR,pope,hallu,yue2024mmmu,tong2024eyes,hiippala2021ai2d,realworldqa,textvqa,ning2025video,A2026}, establishing them as a critical step towards AGI.


\begin{figure}[t]
    \centering
\includegraphics[width=.7\columnwidth]{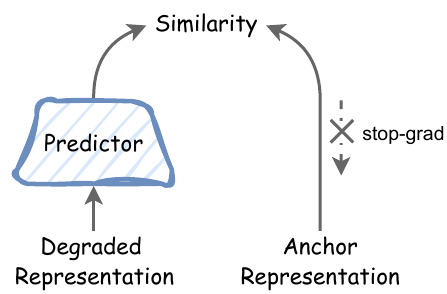}
    \vspace{-2mm}
    \caption{
    \textbf{The Proposed PRe Framework.}
    PRe enforces the degraded representations to predict their initial, clean anchor representations via a lightweight prediction head.
}
    \label{fig:teaser}
    \vspace{-6mm}
\end{figure}

The dominant architectural paradigm for these MLLMs involves a straightforward yet effective design: a powerful, pre-trained vision encoder~\cite{jiang2023clip,tschannen2025siglip} is aligned with an LLM via a lightweight projector that maps visual features into the language embedding space.  The entire system is then fine-tuned almost exclusively through a language-centric objective, \ie, next-token prediction on text data. 
While this approach has proven successful in answering visual questions, it establishes a fundamental asymmetry in the optimization process.   The visual representations are not optimized for their own fidelity; rather, they are implicitly and progressively transformed with the singular goal of better serving the final text generation task.   
Several prior works~\cite{chen2024image,zhang2025cross,jiang2025devils,kang2025see,neo2024towards,wang2025towards,kaduri2025s} have provided valuable insights into this internal process, often utilizing attention analysis or probing techniques to trace how visual information flows and contributes to the model's textual output. However, these existing analyses primarily focus on the cross-modal functionality of visual features, \ie, how well they serve the language task. Their inherent capacity to support fundamental visual tasks independent of language remains largely unexplored. This raises a critical, yet largely overlooked question: \textbf{What is the cost of this purely language-driven training to the visual representations themselves?}
Answering this question is critical as robust and reliable MLLMs should be both eloquent communicators (good at answering visual questions) and sharp-eyed observers (have a robust visual foundation).


To this end, we conduct a comprehensive diagnostic analysis to systematically investigate this cost.  Our investigation begins by directly measuring the functional validity of visual representations at each layer of various MLLMs on standard visual classification tasks.
As shown in~\cref{fig:linear_probe_main}, our analysis uncovers a striking and robust pattern of \textbf{global functional degradation}. Compared to the initial visual features fed into the LLM, the representations in the intermediate layers exhibit a significant degradation in their classification capability.  
Recent studies~\cite{jiang2025devils, zhang2025cross} have highlighted the critical importance of intermediate-layer visual representations in MLLMs,  suggesting they encode rich, abstract semantics crucial for the model's answering. This raises a problem,
how can these layers be simultaneously so crucial for final answers, yet so functionally degraded for visual tasks?
To unravel this paradox, we shift our analysis to the microscopic patch level. As shown in~\cref{fig:patch_contrast} and~\cref{fig:patch_similarity_heatmap}, compared to the initial visual features, the representations processed by the LLM exhibit a pronounced blurring of semantic boundaries between different objects. For example, the semantic contrast ratio consistently declines through the intermediate layers, and visualizations confirm that similarity from a single patch spills over to unrelated objects. 
This \textbf{patch structure degradation} is the direct microscopic cause of the macroscopic functional degradation. Considering that the model progressively aligns visual features with text, we argue that \textbf{the visual degradation is sacrificed for achieving advanced language capabilities}. The model fuses local semantics to build a globally coherent, abstract representation suitable for complex linguistic description, thereby sacrificing the fine-grained, visually-discriminative structure of the initial representation to a certain extent. The statistical properties~\cref{fig:statistical_properties} and the dynamics of the model's visual and linguistic capabilities during pre-training~\cref{fig:tradeoff_evolution} further substantiate this hypothesis.

 Considering that visual ability is the cornerstone of MLLMs, we argue that a robust MLLM should achieve high-level semantic abstraction while simultaneously preserving core visual fidelity.
 Inspired by the principles of predictive coding~\cite{rao1999predictive,friston2005theory,assran2023self}, which posit that an efficient neural system constantly predicts its own lower-level signals to maintain a coherent world model, we propose \textbf{PRe (Predictive Regularization)} to regularize the degradation.
As shown in~\cref{fig:teaser}, PRe compels the degraded visual representations of the intermediate layers to predict the initial anchor features. This predictive objective, optimized concurrently with the standard text generation loss, encourages the model to enhance its linguistic capabilities without sacrificing its foundational visual fidelity. Extensive experiments across vision-language tasks confirm that mitigating this visual degradation effectively boosts performance, underscoring the critical importance of fostering robust internal visual representations within MLLMs for comprehensive multimodal understanding.

Our main contributions are threefold:
\begin{itemize}
    \item We are the first to systematically diagnose the visual degradation phenomenon, providing a multi-level explanation for this degradation, linking the global functional degradation to a microscopic patch degradation.
    \item We identify the degradation as a visual sacrifice for serving the language objective, and propose PRe to counteract this degradation.
    \item  Extensive experiments demonstrate the effectiveness of PRe, emphasizing the significance of robust internal visual representations for MLLMs.
\end{itemize}
\section{Related Work}
\label{sec:related}

\paragraph{Multimodal Large Language Models (MLLMs).}
 Large Language Models (LLMs)~\cite{devlin2019bert,radford2018improving,radford2019language,touvron2023llama,bai2023qwen,2023internlm,liu2024deepseek} have achieved significant success in natural language processing. Building on this, researchers have created Multimodal Large Language Models (MLLMs)~\cite{liu2024llavanext, liu2023improvedllava, chen2024internvl,wang2025internvl3_5,lu2024deepseekvl,Qwen-VL,Qwen2.5-VL} by incorporating visual information, aiming to bridge the vision-language semantic gap and process diverse multimodal data. 
 Current research on MLLMs focuses on optimizing visual encoders for high-quality feature extraction~\cite{tong2024eyes, jiang2023clip, shi2024eagle, fan2025scaling, shen2024mome, wu2025tove}; enhancing precise visual perception and fine-grained comprehension of image content to mitigate multimodal hallucinations~\cite{jiang2025devils, kang2025see,zhou2023analyzing,huo2024self,leng2024mitigating,liu2024paying} for accurate visual QA; and improving multimodal instruction following and reasoning, bolstering the model's ability to execute complex vision-language tasks through Chain-of-Thought~\cite{wei2022chain, zhang2023multimodal, wang2025multimodal, xu2025llava, wang2024t,yao2024promptcot} and reinforcement learning strategies~\cite{guo2025deepseek,ouyang2022training,shao2024deepseekmath,schulman2017proximal,liu2025visual,shen2025vlm,zheng2025deepeyes,litempsamp,zeng2025glimpse} like PPO and GRPO.
 These works primarily emphasize the effective utilization of visual information within textual contexts, aiming to optimize vision-language alignment for language tasks. However, they often overlook the intrinsic integrity and robustness of MLLM's internal visual features. While some studies introduce auxiliary visual tasks to MLLMs, such as image reconstruction~\cite{wang2024ross}, masking visual tokens for redundancy reduction~\cite{xie2024victor}, or aligning visual hidden states with foundation models~\cite{yoon2025viral}, these efforts differ from our focus on understanding and mitigating representational degradation. In this paper, we first analyze the visual representation degradation problem within MLLMs, a critical aspect overlooked by prior work. Subsequently, our approach aims to directly maintain the visual semantic integrity and stability of MLLM's internal visual features through a self-supervised consistency learning paradigm, specifically addressing the potential degradation of visual representations during complex cross-modal interactions.

\myPara{Visual Representation Learning.}
Learning robust and generalizable visual representations is a core objective in computer vision. The dominant paradigm has shifted from early supervised learning approaches~\cite{deng2009imagenet,krizhevsky2012imagenet,he2016deep} to Self-Supervised Learning (SSL)~\cite{liu2021self,jaiswal2020survey,jing2020self}, which aims to derive features from vast quantities of unlabeled data. SSL has progressed along two main technical trajectories: contrastive learning~\cite{hadsell2006dimensionality,bachman2019learning,chen2020simple,he2020momentum,hjelmlearning,henaff2020data,misra2020self,oord2018representation,wu2018unsupervised,wang2025get} and generative learning~\cite{baobeit,he2022masked,zhouimage,xie2022simmim,wei2022masked,assran2023self}.
Contrastive learning operates by learning a representation space where positive samples are pulled together, and negative samples are pushed apart to avoid trivial solutions. Seminal works like SimCLR~\cite{chen2020simple} and MoCo~\cite{he2020momentum} achieve this by relying on a large number of negative samples or a momentum encoder. 
Subsequent methods like SwAV~\cite{caron2020unsupervised}, BYOL~\cite{grill2020bootstrap}, SimSiam~\cite{chen2021exploring}, DINO~\cite{caron2021emerging}, and Barlow Twins~\cite{zbontar2021barlow} successfully obviated the need for negative samples through diverse strategies such as online clustering, asymmetric prediction, self-distillation, and feature decorrelation.
Generative learning trains models by predicting masked or missing content.
This can be done at the pixel level, as seen in MAE~\cite{he2022masked}, SimMIM~\cite{xie2022simmim}. More recent architectures, such as the JEPA~\cite{assran2023self}, have shifted this task to a latent space, predicting the representations of masked content.
A unifying principle connecting many of these advanced SSL methods is Predictive Coding~\cite{friston2005theory,rao1999predictive}, a theory from neuroscience suggesting that learning occurs by predicting internal states or sensory inputs.  
This predictive mechanism is explicit in generative approaches like JEPA and implicitly implemented in contrastive methods like SimSiam via its asymmetric predictor. 
 Our work is directly inspired by this powerful predictive principle.    
 However, we fundamentally re-contextualize this mechanism: instead of using it as a pretext task to learn representations in a unimodal setting, we employ it as a regularizer to preserve the fidelity of an existing visual representation within a multimodal model against the degenerative pressure of the text-alignment objective.

\section{Preliminaries}

In this section, we formalize the predominant architecture of MLLMs and delineate their standard training paradigm. 

\myPara{The Architecture of MLLMs.} A typical MLLM integrates a pre-trained vision encoder with a large language model. Given an input image $\mathbf{I} \in \mathbb{R}^{H \times W \times C}$, a vision encoder $f_v(\cdot)$ first processes it into a sequence of $N_p$ patch features $\mathbf{Z} = \{\mathbf{z}_1, \dots, \mathbf{z}_{N_p}\}$, where each $\mathbf{z}_i \in \mathbb{R}^{d_v}$. This sequence constitutes the initial visual representation. After that, a projection layer $f_p(\cdot)$ is employed to map these features into the language embedding space, yielding a sequence of projected visual tokens $\mathbf{H}_v^0 = \{\mathbf{h}_{v,1}^0, \dots, \mathbf{h}_{v,{N_p}}^0\}$, where each $\mathbf{h}_{v,i}^0 \in \mathbb{R}^{d_l}$. These visual tokens are then concatenated with the text token embeddings $\mathbf{H}_t$ derived from a text prompt $\mathbf{T}_{\text{prompt}}$. The resulting mixed sequence, $\mathbf{H}_{\text{in}} = [\mathbf{H}_t, \mathbf{H}_v^0]$, is fed into the Large Language Model $f_l(\cdot)$, which is an auto-regressive Transformer decoder. The LLM processes this sequence through its multiple self-attention layers, and we denote the output hidden states at layer $l$ as $\mathbf{H}^l = \{\mathbf{H}_t^l, \mathbf{H}_v^l\}$.

\myPara{The Training Paradigm.} The training of MLLMs is exclusively driven by a language modeling objective. Given a dataset of image-text pairs $(\mathbf{I}, \mathbf{T})$, where $\mathbf{T}$ is the desired response, the model is trained to maximize the conditional probability of generating the answer text $\mathbf{T}_{\text{answer}} = \{y_1, \dots, y_K\}$ given the image $\mathbf{I}$ and a prompt $\mathbf{T}_{\text{prompt}}$. This is achieved by minimizing the standard auto-regressive loss, also known as the next-token prediction loss:
\begin{equation}
\label{eq:lm_loss}
\mathcal{L}_{\text{LM}} = - \sum_{k=1}^{K} \log P(y_k | \mathbf{I}, \mathbf{T}_{\text{prompt}}, y_{<k}; \Theta)
\end{equation}
where $P(y_k | \cdot)$ is the model's predicted probability for the correct token at step $k$. 
\section{PRe: Predictive Regularization to Mitigate Representation Degradation}
\label{sec:method}

In this section, we first delve into a critical yet under-explored issue within MLLMs: \textbf{the degradation of visual representations}. Through a series of diagnostic analyses in~\cref{subsec:VRD}, we reveals that this degradation manifests as both a drop in global representation performance and a dissociation of local semantic structures at the patch level.
 We then propose \textbf{PRe (Predictive Regularization)} in Section~\ref{subsec:PR}, a novel and lightweight method specifically designed to counteract this degradation by preserving the fidelity of the initial visual representations.

\subsection{Visual Representation Degradation}
\label{subsec:VRD}

Understanding the internal evolutionary trajectory of visual representations within MLLMs is crucial for diagnosing model behaviors, interpreting their capabilities, and guiding future architectural improvements. As visual features traverse the deep layers of a language model, they are progressively transformed to better serve the final text generation objective. 
Several prior works~\cite{chen2024image,zhang2025cross,jiang2025devils,kang2025see,neo2024towards,wang2025towards,kaduri2025s} have provided valuable insights into this process.
However, these existing analytical approaches primarily focus on the {cross-modal functionality} of visual features. The {intrinsic visual fidelity} of these representations remains largely unexplored. 
This leaves a critical question unanswered: What is the cost of this deep, language-driven alignment?  Does the process of refining visual features for language generation simultaneously degrade their fundamental visual capabilities? 
To investigate this, we conduct a multi-faceted analysis, starting from the functional validity of global representations down to the semantic structure at the patch level.

\begin{figure}[t]
    \centering
\includegraphics[width=\columnwidth]{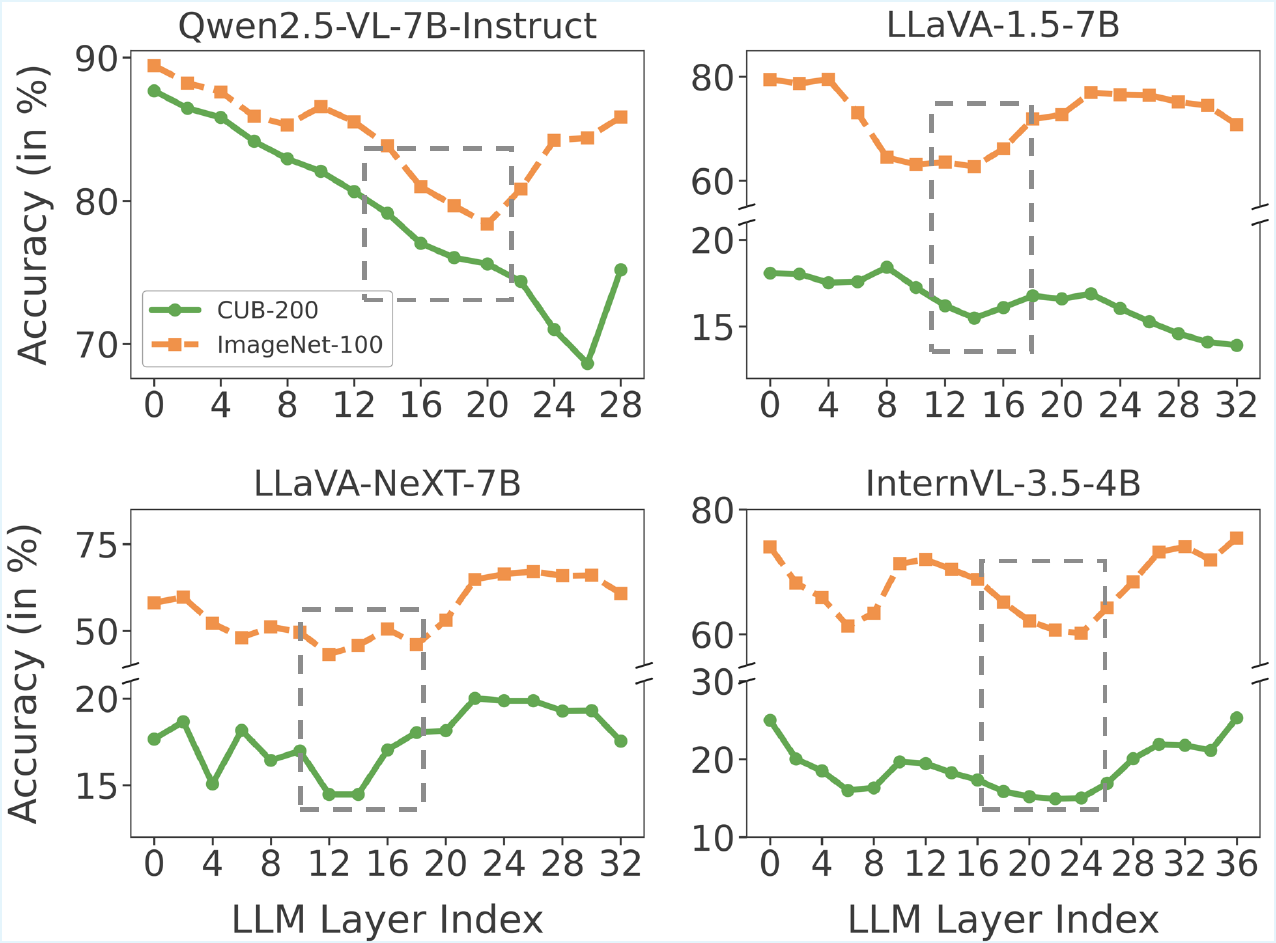}
    \vspace{-6mm}
    \caption{\textbf{Linear probe results on global visual features.} Relative to the initial representation (Layer 0), a consistent performance drop is observed in the intermediate layers.
}
    \label{fig:linear_probe_main}
    \vspace{-6mm}
\end{figure}

\myPara{Global Functional Degradation.}
To probe the functional validity of visual representations, we perform extensive linear probing experiments across multiple datasets (CUB-200~\cite{WahCUB_200_2011}, ImageNet-100~\cite{krizhevsky2012imagenet}) and MLLM architectures~\cite{xu2025llava,liu2024llavanext,wang2025internvl3_5,Qwen2.5-VL}. 
 For each layer $l$, we first extract the  global aggregated  feature $\bar{\mathbf{h}}^l = \frac{1}{{N_p}} \sum_{i=1}^{{N_p}} \mathbf{h}_{v,i}^l$,
and then perform linear probing~\cite{he2022masked} by training a linear classifier on these frozen features. 
As shown in ~\cref{fig:linear_probe_main},  the results reveal a critical and remarkably robust phenomenon:  there exists a significant performance drop in the intermediate layers relative to the initial, high-fidelity representation at Layer 0. 
The fact that this global functional degradation occurs robustly across different models and datasets confirms it is a systematic byproduct of the prevailing training paradigm, which is singularly focused on language generation, rather than a coincidental artifact.

\begin{figure}[t]
    \centering
\includegraphics[width=\columnwidth]{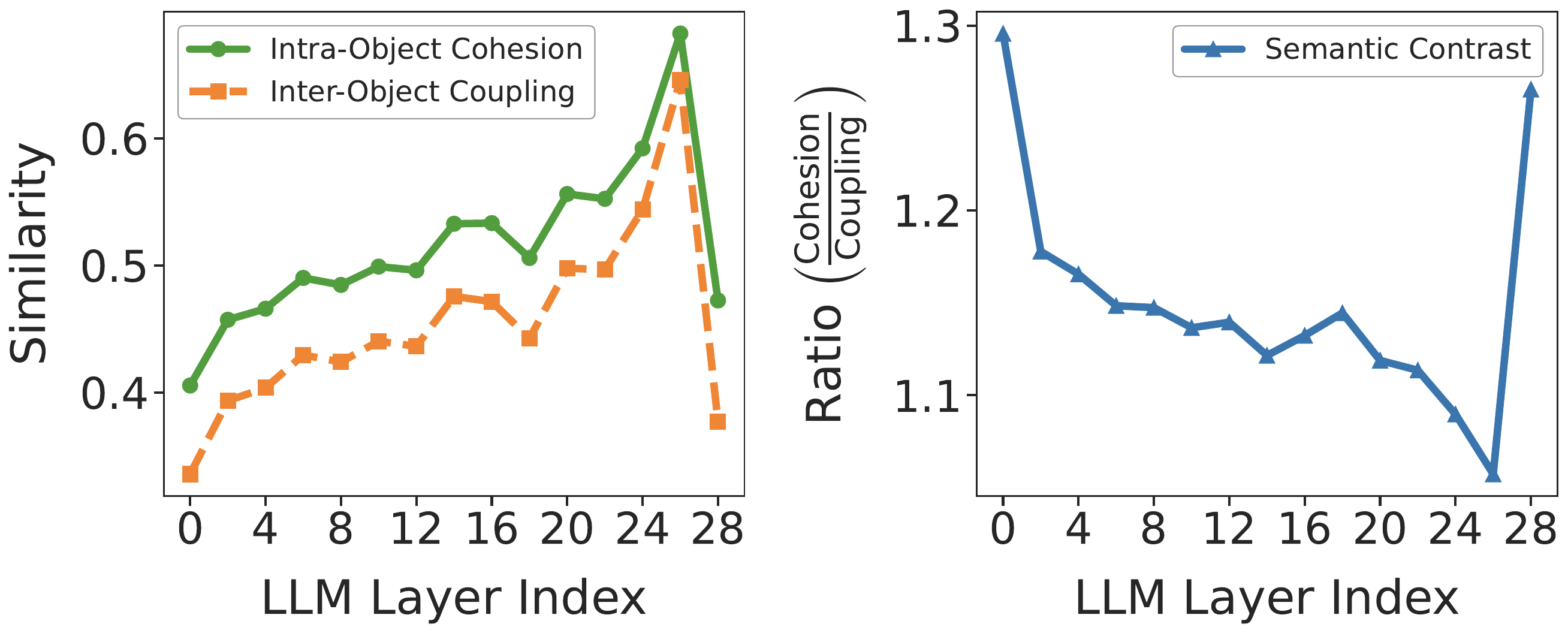}
    \vspace{-6mm}
    \caption{
    \textbf{Evolution of patch-level semantics.}
The intra- and inter-object similarities rise through most layers, leading to reduce the semantic contrast ratio, revealing patch structure degradation.
    }
    \label{fig:patch_contrast}
    \vspace{-2mm}
\end{figure}

\begin{figure}[t]
    \centering
\includegraphics[width=\columnwidth]{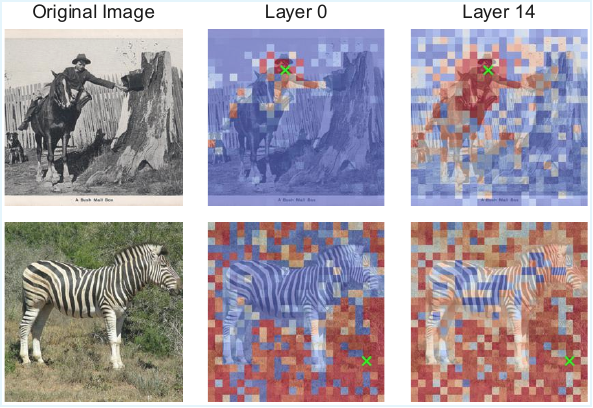}
    \vspace{-6mm}
    \caption{\textbf{The cosine similarity between the patch noted in green and all other patches.} Compared with the initial representation, the middle layer blurs the visual semantic boundaries. (Images are from opensource COCO-stuff~\cite{caesar2018cocostuffthingstuffclasses} dataset)
    }
    \label{fig:patch_similarity_heatmap}
    \vspace{-5mm}
\end{figure}

\myPara{Patch Structure Degradation .}
To understand the microscopic mechanism behind the global functional degradation, we analyze the semantic structure at the patch level. Using segmentation masks from the COCO-stuff dataset~\cite{caesar2018cocostuffthingstuffclasses}, we quantify the separability of patch representations for each layer. Specifically, we measure two metrics: (1) Intra-Object Cohesion, the average cosine similarity between patches belonging to the same object instance, and (2) Inter-Object Coupling, the average similarity between patches from different object instances. We then define the Semantic Contrast ratio as the ratio of Cohesion to Coupling.
Our results, shown in ~\cref{fig:patch_contrast}, reveal that the global functional drop is mirrored by a degradation in patch-level separability. While Cohesion and Coupling both rise through the intermediate layers, Coupling increases at a faster rate, leading to a consistent decrease in the overall Semantic Contrast. \cref{fig:patch_similarity_heatmap} further shows the similarity map from a randomly selected patch to all other patches in an image, for the initial and intermediate layers. In the initial layer, similarity is high almost exclusively for patches belonging to the same object as the probe. In contrast, for the intermediate layer, high similarity spills over to patches of different objects and the background, visually confirming the semantic fusion process. This convergence of findings provides strong evidence that the degradation of macroscopic functionality is rooted in the progressive blurring of semantic boundaries at the microscopic patch level.

\myPara{Degradation as a Visual Sacrifice for Language Capability.}
Our analysis so far has established that mid-layer visual representation degradation is a real and robust phenomenon, mechanistically rooted in patch-level semantic fusion. But what drives the model to adopt this seemingly detrimental strategy? We think this degradation is a visual sacrifice optimized for the model's ultimate objective: complex language generation. 

\begin{figure}[t]
    \centering
\includegraphics[width=\columnwidth]{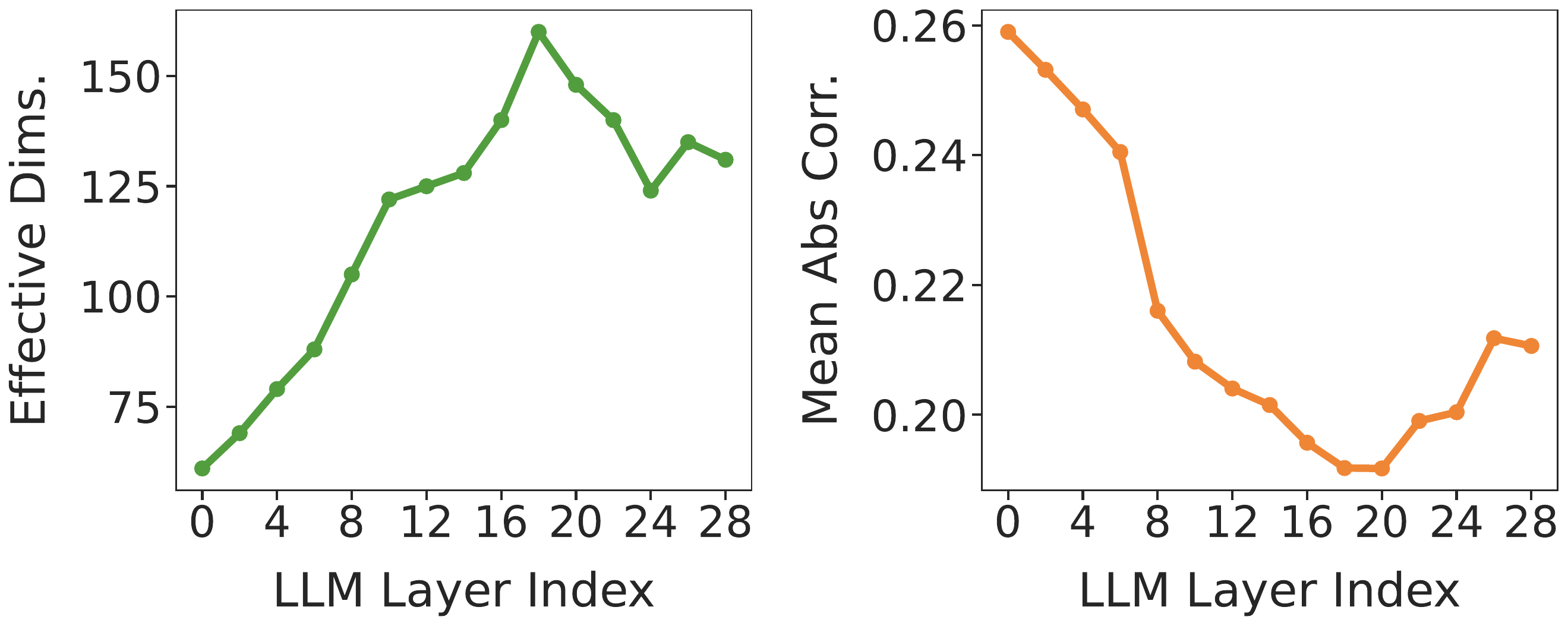}
    \vspace{-6mm}
    \caption{\textbf{The statistical properties of the global representations.}  The increased PCA effective dimension and reduced mean off-diagonal correlation indicate that the representations become both more geometrically complex and statistically independent.
    }
    \label{fig:statistical_properties}
    \vspace{-8mm}
\end{figure}

To investigate this, we first analyze the statistical properties of the global representations $\bar{\mathbf{h}}^l$. We measure the \textbf{geometric complexity} via the number of principal components needed to explain 95\% of the variance (PCA effective dimension), and the \textbf{statistical independence} via the mean absolute off-diagonal correlation.   As shown in~\cref{fig:statistical_properties}. We find that the intermediate layers, where visual representation degrades, counter-intuitively exhibit better statistical structure: their geometric complexity peaks, indicating the highest information capacity, while their feature correlation drops to a minimum, indicating the best statistical independence. We think that this sophisticated structural reshaping is a deliberate strategy to build a powerful and flexible ``workspace" for language. 
For instance, consider an image containing a ``red wooden chair'', the initial visual representation might encode ``red'', ``wooden'', and ``chair'' in a compressed and entangled manner. To generate a descriptive sentence, LLM must perform a {disentanglement and expansion}, thus inherently increasing the representation's complexity and statistical independence.
The visual degradation, therefore, is the direct cost of this transformation, where the simple linear structure is sacrificed to enable more complex linguistic capabilities. 
This aligns with findings from prior work, which suggest that as information propagates through the layers of a large model, representations progressively shift from encoding low-level perceptual features to high-level conceptual semantics~\cite{wang2025towards, neo2024towards,jiang2025devils}. 

\begin{figure}[t]
    \centering
\includegraphics[width=.66\columnwidth]{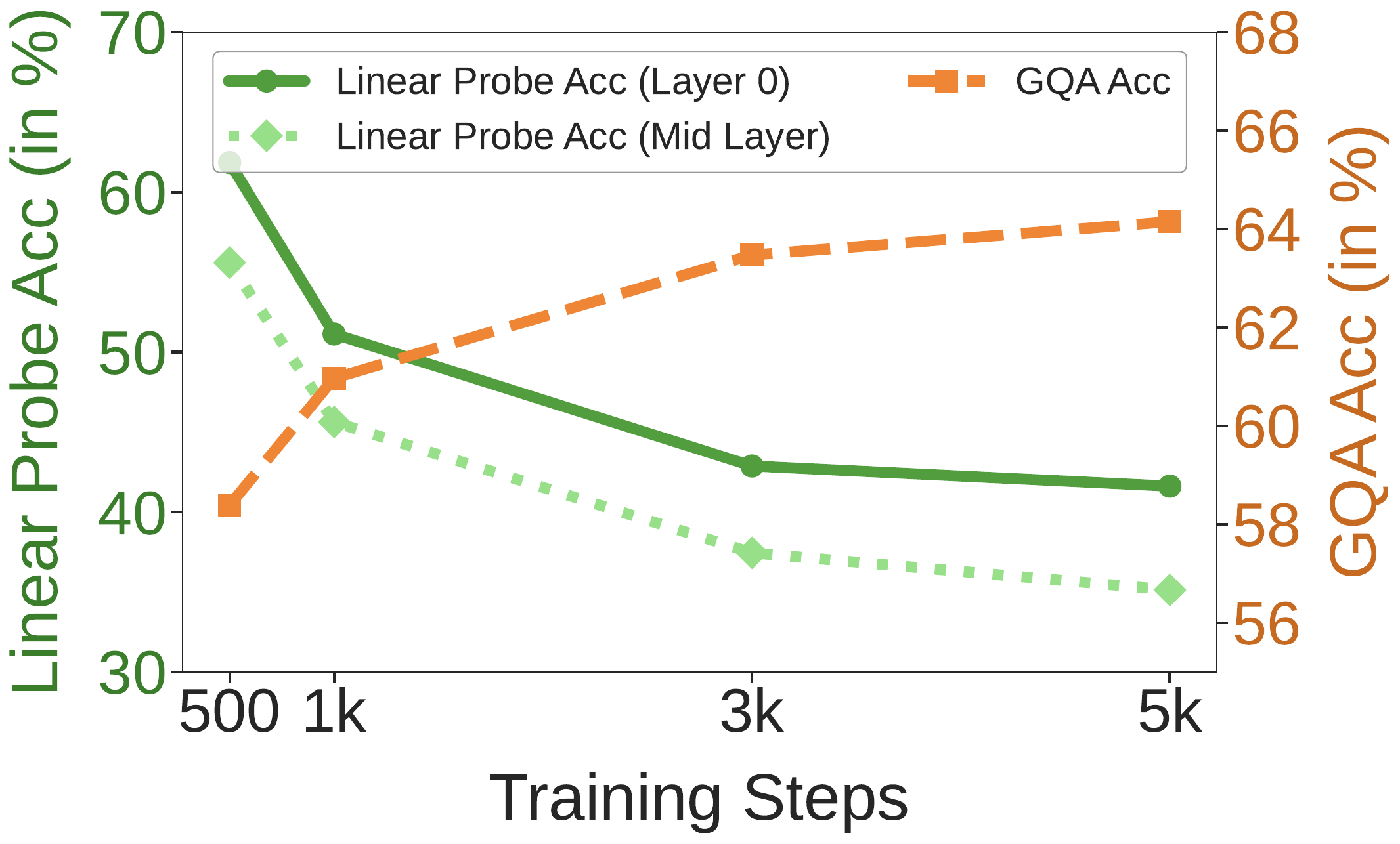}
    \vspace{-2mm}
    \caption{\textbf{Visual and language capabilities during pre-training.} {Results show that the language-driven optimization degrades intrinsic visual fidelity.}
    }
    \label{fig:tradeoff_evolution}
    \vspace{-3mm}
\end{figure}

Furthermore, we confirm that this sacrifice is an inherent byproduct of the language-driven training by tracing the dynamic evolution of the model's visual and linguistic capabilities throughout its pre-training process. As highlighted in~\cref{fig:tradeoff_evolution}, a clear inverse correlation emerges: as the model's visual question answering performance (\eg, on GQA~\cite{Hudson_2019_CVPR}) steadily improves with more training steps, its intrinsic visual fidelity, measured by linear probe accuracy on the same checkpoints, simultaneously and continuously degrades.  This dynamic demonstrates that the degradation of visual representation is not a random artifact, but an inherent, systemic byproduct of an architecture singularly optimized for language generation, in the complete absence of a direct visual objective.

\subsection{Predictive Regularization}
\label{subsec:PR}
In~\cref{subsec:VRD}, we identify a critical visual representation degradation phenomenon, which is a sacrifice to construct representations for complex language generation, and is a direct consequence of the architecture being designed for a single optimized text objective in the absence of a visual supervision signal. The harshness of the visual degradation may lead to the irreversible loss of critical visual details, creating a bottleneck that limits the model's ultimate performance on nuanced vision-language tasks requiring high fidelity.

A more ideal trajectory would achieve semantic abstraction more gracefully, without such a drastic dip in functional validity. This motivates our work: can we regularize this inherent evolutionary path to make it more efficient and less destructive? To this end, we propose \textbf{PRe (Predictive Regularization)},  a method designed to bridge the gap between visual perception and high-level language conception by establishing a {visual-preserving shortcut}.

As shown in~\cref{fig:teaser}, the core idea of PRe is straightforward yet powerful: we enforce that the degraded visual representations in the intermediate layers must retain sufficient information to reconstruct their initial, high-fidelity state. 
For a given image, the patch features $\mathbf{H}_v^0$ that are fed to the LLM serve as a stable, detached anchor via a {stop-gradient} operation. Concurrently, the degraded visual hidden states $\mathbf{H}_v^l$ from an intermediate LLM layer are passed through a shallow prediction head $f_{pred}(\cdot)$, a 2-layer MLP.
The objective of PRe is to minimize the negative cosine similarity between the predicted features and the anchor features:
\begin{equation}
    \mathcal{L}_{\text{PRe}} = - \frac{1}{N_p} \sum_{i=1}^{N_p} \mathcal{D}\left(f_{pred}(\mathbf{h}_{v,i}^l),  \text{stopgrad}(\mathbf{h}_{v,i}^0) \right)
    \label{eq:pre_loss}
\end{equation}
where $\mathcal{D}(p, z) = (p/\|p\|_2) \cdot (z/\|z\|_2)$ denotes the cosine similarity. 
Our initial design aligns the LLM's intermediate visual features with its input visual features. A significant benefit of this approach is its broad applicability to diverse MLLMs, which we further investigate in~\cref{subsec:exp}.
The final training objective is a weighted sum of the original language modeling loss $\mathcal{L}_{\text{LM}}$ (\cref{eq:lm_loss}) and our proposed regularization loss: $\mathcal{L}_{\text{total}} = \mathcal{L}_{\text{LM}} + \lambda \mathcal{L}_{\text{PRe}}$, where $\lambda$ is a balancing hyperparameter.

The principle of PRe is deeply inspired by theories of predictive coding~\cite{rao1999predictive,friston2005theory}, which posit that an efficient neural system constantly predicts its own lower-level signals from higher-level representations. 
By re-contextualizing this predictive principle as a regularizer, PRe aims to anchor the abstracting representations to their perceptual origins, thereby counteracting the excessive semantic fusion and preserving local fidelity.

\section{Experiments}
\label{sec:exp}
\begin{figure}[t]
    \centering
\includegraphics[width=\columnwidth]{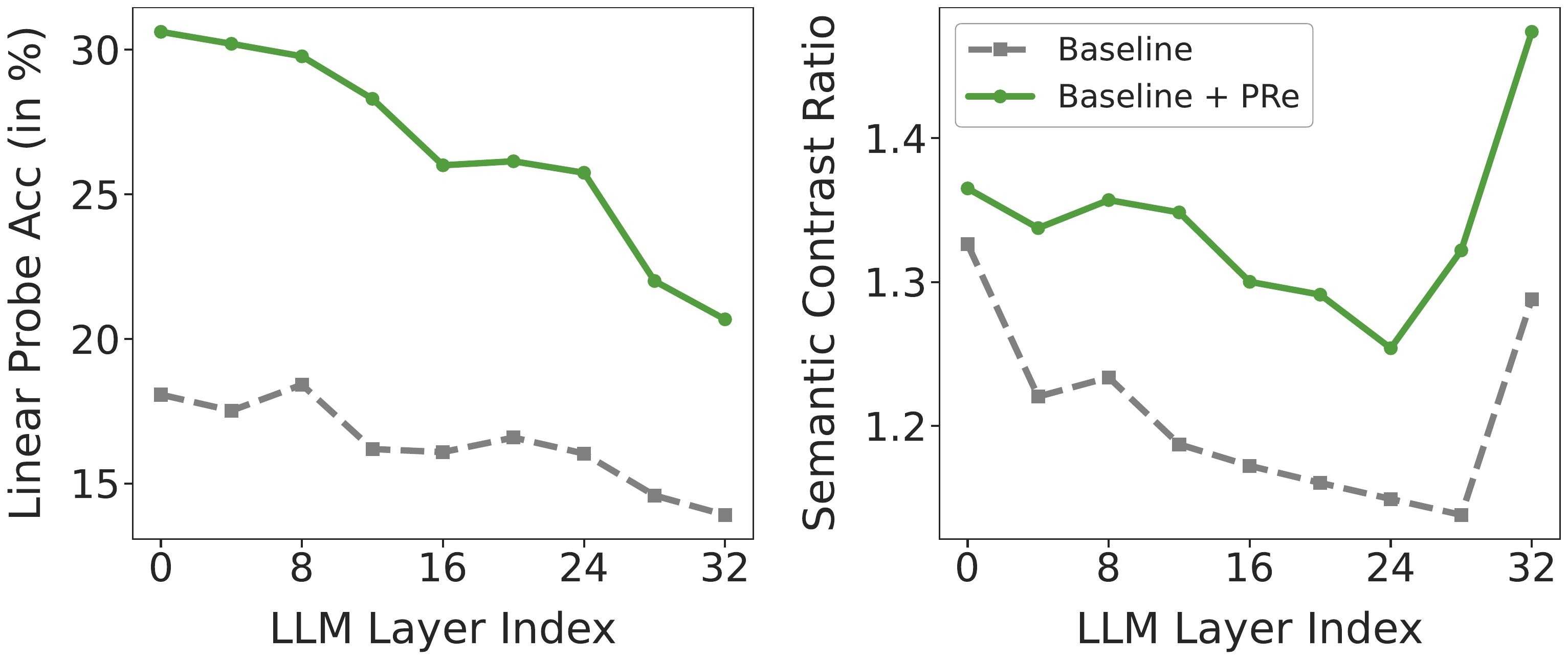}
    \vspace{-6mm}
\caption{\textbf{The effectiveness of PRe in enhancing intrinsic visual capabilities of LLMs.} Baseline architecture using CLIP-ViT-L/14@336 (frozen) and Vicuna-7B-v1.5.
    }
    \label{fig:pre_vs_baseline_comparison}
    \vspace{-5mm}
\end{figure}

Our primary aim in this work is not to establish a new state-of-the-art MLLM, but rather to investigate the phenomenon of visual representation degradation and explore the potential of improving MLLM performance through intrinsic visual representation optimization. Consequently, our experiments are designed to answer the following questions:
\begin{enumerate}
    \item \textbf{Effectiveness on Language Tasks:} Does mitigating the identified visual degradation (via PRe) translate to tangible improvements on downstream vision-language tasks? (\cref{fig:pre_vs_baseline_comparison},~\cref{fig:similarity_heatmap_comparison}, and~\cref{tab:main_results})
    \item \textbf{Impact of Design Choices:} How to select the degradation representation and the anchor representation? (\cref{tab:ablation_layer_choice},~\cref{fig:token_prediction_stacked_bar},~\cref{tab:Anchor_Source},~\cref{fig:f_ablation}, and~\cref{fig:lam_ablation})
\end{enumerate}

\subsection{Setup}
Our training follows the LLaVA series~\cite{xu2025llava,liu2024llavanext}. 
The training data are LLaVA-558K and LLaVA-665K for the pre-training stage and the instruction tuning stage, respectively. We evaluate PRe across multiple vision encoders (CLIP~\cite{jiang2023clip} and SigLIP~\cite{tschannen2025siglip}), LLMs (Vicuna~\cite{vicuna} and Qwen~\cite{yang2025qwen3}) and training strategies. $\lambda$ is set to 0.5, and the degraded layer in~\cref{eq:pre_loss} is set to the middle of the LLM (16 for Vicuna, 14 for Qwen). For evaluation, we assess the performance of our method across multiple benchmarks, including GQA~\cite{Hudson_2019_CVPR}, MMMU~\cite{yue2024mmmu}, RealWorldQA~\cite{realworldqa}, TextVQA~\cite{textvqa}, AI2D~\cite{hiippala2021ai2d}, MMStar~\cite{mmstar}, ScienceQA~\cite{sqa},  OCRbench~\cite{liu2023ocrbench} and MMVP~\cite{tong2024eyes}. More details can be found in the supplementary material (\textit{supp}).

\subsection{Experiments and Analysis}
\label{subsec:exp}
\begin{figure}[t]
    \centering
\includegraphics[width=\columnwidth]{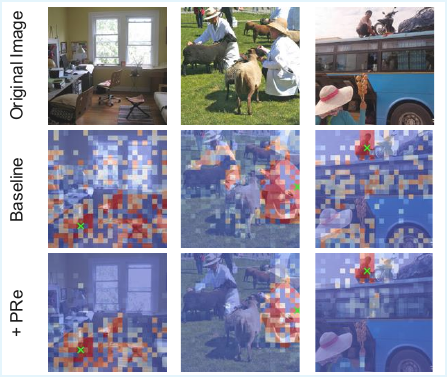}
    \vspace{-6mm}
    \caption{\textbf{The cosine similarity between the patch noted in green and all other patches in the middle layer.} Images are from opensource COCO-stuff~\cite{caesar2018cocostuffthingstuffclasses} dataset. 
    }
    \label{fig:similarity_heatmap_comparison}
    \vspace{-6mm}
    
\end{figure}

\myPara{PRe Enhances Intrinsic Visual Capabilities of LLMs.}
To start with, we investigate whether PRe can effectively enhance the intrinsic visual capabilities of the LLM by mitigating the degradation identified in~\cref{subsec:VRD}. As shown in Figure~\ref{fig:pre_vs_baseline_comparison}, when PRe is applied, the model's internal representations exhibit substantial improvements across all layers. At the macroscopic level, the linear probe performance of global features is significantly boosted compared to the baseline, indicating that the self-prediction objective successfully enforces the preservation of linearly separable visual information. Furthermore, at the microscopic level, the patch-level semantic contrast ratio is also consistently elevated across all layers, with the degradation relative to the initial layer being markedly reduced. \cref{fig:similarity_heatmap_comparison} provides a qualitative illustration of this effect, showing that PRe significantly suppresses the semantic fusion between patches of different objects compared to the baseline.

\begin{table*}[t!]
    \centering\small
    \caption{
    \textbf{The effectiveness of the predictive regularization among various LLMs and visual encoders}. * indicates the visual encoder is frozen during training, without * indicates the visual encoder is trainable.
    }
    \label{tab:main_results}
    \vspace{-5pt}
    \setlength{\tabcolsep}{2pt}
    \resizebox{\textwidth}{!}{%
    \begin{tabular}{lc llllllllll}
    \toprule
    \multirow{2}{*}{Visual Encoder} & \multirow{2}{*}{$\mathcal{L}_{\text{PRe}}$ }& \multicolumn{5}{c}{General \& Knowledge} & \multicolumn{2}{c}{OCR-related} & \multicolumn{2}{c}{Vision-centric} \\ 
\cmidrule(rl){3-7} 
\cmidrule(rl){8-9}
\cmidrule(rl){10-11} & & GQA & MMMU & AI2D & MMStar & SQA$^\text{I}$ & VQA$^\text{T}$ & OCRbench & RWQA & MMVP\\
\midrule
\rowcolor{lightgray}
    
    \multicolumn{11}{l}{\textit{LLM: Vicuna-7B-v1.5 }} \\

        \multirow{2}{*}{CLIP-ViT-L/14@336 * } & -- & 62.0&	35.7 &55.4 &	30.3  &66.8	&45.5	&	318 &54.8 &20.0 \\
    & \checkmark & \greenscore{62.7}{+0.7} &  
    \greenscore{36.1}{+0.4}&\greenscore{57.1}{+1.7} & \greenscore{34.6}{+4.3}&\greenscore{67.5}{+0.7} &\greenscore{46.6}{+1.1}& \greenscore{329}{+11} &
    \greenscore{55.4}{+0.6}& \greenscore{22.0}{+2.0}\\
      \multirow{2}{*}{SigLIP2-ViT-SO400M/14@384 *} & -- & 62.5	&36.5 	& 58.8&	36.8 &68.4 &	57.9	& 396&56.9	& 35.3 \\
    & \checkmark &  \greenscore{64.3}{+1.8}	& \redscore{36.4}{-0.1}	& \greenscore{59.3}{+0.5}	& \redscore{36.7}{-0.1}	& 	\greenscore{70.0}{+1.6}		& \greenscore{59.2}{+1.3}	& \greenscore{401}{+5} &  \greenscore{57.1}{+0.2} & \greenscore{36.7}{+1.4}\\
    \multirow{2}{*}{SigLIP2-ViT-SO400M/14@384} & -- & 64.7 & 36.2 & 59.7 & 36.3 & 71.0 & 58.8 & 409 & 56.8 & 41.3 \\
& \checkmark & \greenscore{64.9}{+0.2} &  \greenscore{36.5}{+0.3} &  \greenscore{60.1}{+0.4} &  \greenscore{36.3}{+0.0}&  \greenscore{71.4}{+0.4} &  \greenscore{59.5}{+0.7} & \greenscore{415}{+6}& \greenscore{57.0}{+0.2}& \greenscore{44.7}{+3.4}   \\

    \hline
    \rowcolor{lightgray}
    
    \multicolumn{11}{l}{\textit{LLM: Qwen2.5-7B-Instruct }} \\
\multirow{2}{*}{CLIP-ViT-L/14@336 *} & -- &  63.0 	&45.1 &66.1	& 43.5 &75.4		&46.2		&332 &56.9 & 37.3 \\
    & \checkmark & \redscore{62.9}{-0.1}	&\redscore{44.4}{-0.7} &\greenscore{66.3}{+0.2}&	\redscore{42.6}{-0.9}	&	\greenscore{75.7}{+0.3}		&\greenscore{46.3}{+0.1}	&	\greenscore{334}{+2} &\greenscore{59.3}{+2.4} & \greenscore{40.7}{+3.4}
    \\
        \multirow{2}{*}{CLIP-ViT-L/14@336} & -- & 62.7&	44.6&	67.4&	44.1&		76.0		&48.3	& 369 &57.8 & 40.0   \\
    & \checkmark &  \greenscore{63.1}{+0.4} & 	\greenscore{45.4}{+0.8}	& 	\greenscore{68.3}{+0.9} &  	\greenscore{45.4}{+1.3}	& 	\greenscore{76.5}{+0.5} &  \greenscore{49.5}{+1.2} &  	\greenscore{385}{+16} & \greenscore{59.2}{+0.4}	& \redscore{39.3}{-0.7}\\
    \multirow{2}{*}{SigLIP2-ViT-SO400M/14@384} & -- & 63.5 & 45.8 & 68.9 & 48.0 & 78.3& 59.2  & 413  & 60.3 & 46.0\\
    & \checkmark  & \greenscore{64.4}{+0.9} & \greenscore{46.2}{+0.4}&\greenscore{69.5}{+0.6}&\redscore{47.8}{-0.2}&\redscore{77.9}{-0.4}&\greenscore{59.7}{+0.5}&\greenscore{428}{+5}&\greenscore{61.9}{+1.6} &\greenscore{46.7}{+0.7}\\

    \bottomrule
    \end{tabular}}
    \vspace{-10pt}
\end{table*}

\myPara{Results on Diverse Architectures.}
To demonstrate that enhancing the LLM's intrinsic visual capabilities leads to tangible benefits on vision-language tasks, 
we present our results on two popular 7B baseline LLMs: Vicuna-7B-v1.5~\cite{vicuna} and Qwen-2.5-7B-Instruct~\cite{bai2023qwen}, with either a CLIP or a SigLIP vision encoder. As shown in~\cref{tab:main_results}, the application of our PRe method yields performance improvements across most tested architectural configurations.
For instance, when equipping the Vicuna + CLIP* model with PRe, we observe the improvement across all evaluation benchmarks. This directly confirms our central hypothesis that mitigating the visual representation degradation translates to tangible enhancements in linguistic capabilities.
Similar positive trends are robustly observed across other configurations. For example, our method boosts the performance of the Qwen + SigLIP on RealWorldQA from 60.3\% to 61.9\%,  boosts the performance of the Qwen + CLIP on MMStar from 44.1\% to 45.4\%, and TextVQA from 48.3\% to 49.5\%. These observed performance gains provide strong evidence for the efficacy of counteracting visual degradation in applications requiring high visual fidelity and foundational visual understanding.

\begin{table}[t!]
    \centering\small
    \caption{\textbf{Results of different regularized layers.}}
    \label{tab:ablation_layer_choice}
    \vspace{-5pt}
    \setlength{\tabcolsep}{3.7pt}
    \resizebox{\linewidth}{!}{%
    \begin{tabular}{c ccccc}
    \toprule
     Select Layer & GQA & MMMU & VQA$^\text{T}$ & RWQA & MMVP \\
    \hline
    \rowcolor{lightgray}
    \multicolumn{6}{l}{\textit{CLIP (frozen) + Vicuna-7B-v1.5 }} \\
    baseline & 62.0	&35.7 &45.5 &54.8&20.0 \\
    mid-layer & \textbf{62.7 }& \textbf{36.1}& \textbf{46.6} &\textbf{55.4}&22.0\\
    last-layer &62.4 & 35.6 & 45.7 &54.5& \textbf{25.3} \\
    \hline
    \rowcolor{lightgray}
    \multicolumn{6}{l}{\textit{SigLIP2 (trainable)+ Qwen2.5-7B-Instruct }} \\
    baseline & 63.5&45.8&59.2 & 60.3 &46.0\\
    mid-layer & \textbf{64.4}&\textbf{46.2}&\textbf{59.7} & \textbf{61.9} & 46.7\\
    last-layer & 63.2 & 46.1 &58.9 & 61.2 & \textbf{47.3} \\
    \bottomrule
    \end{tabular}}
    \vspace{-4mm}
\end{table}

\begin{figure}[t]
    \centering
\includegraphics[width=\columnwidth]{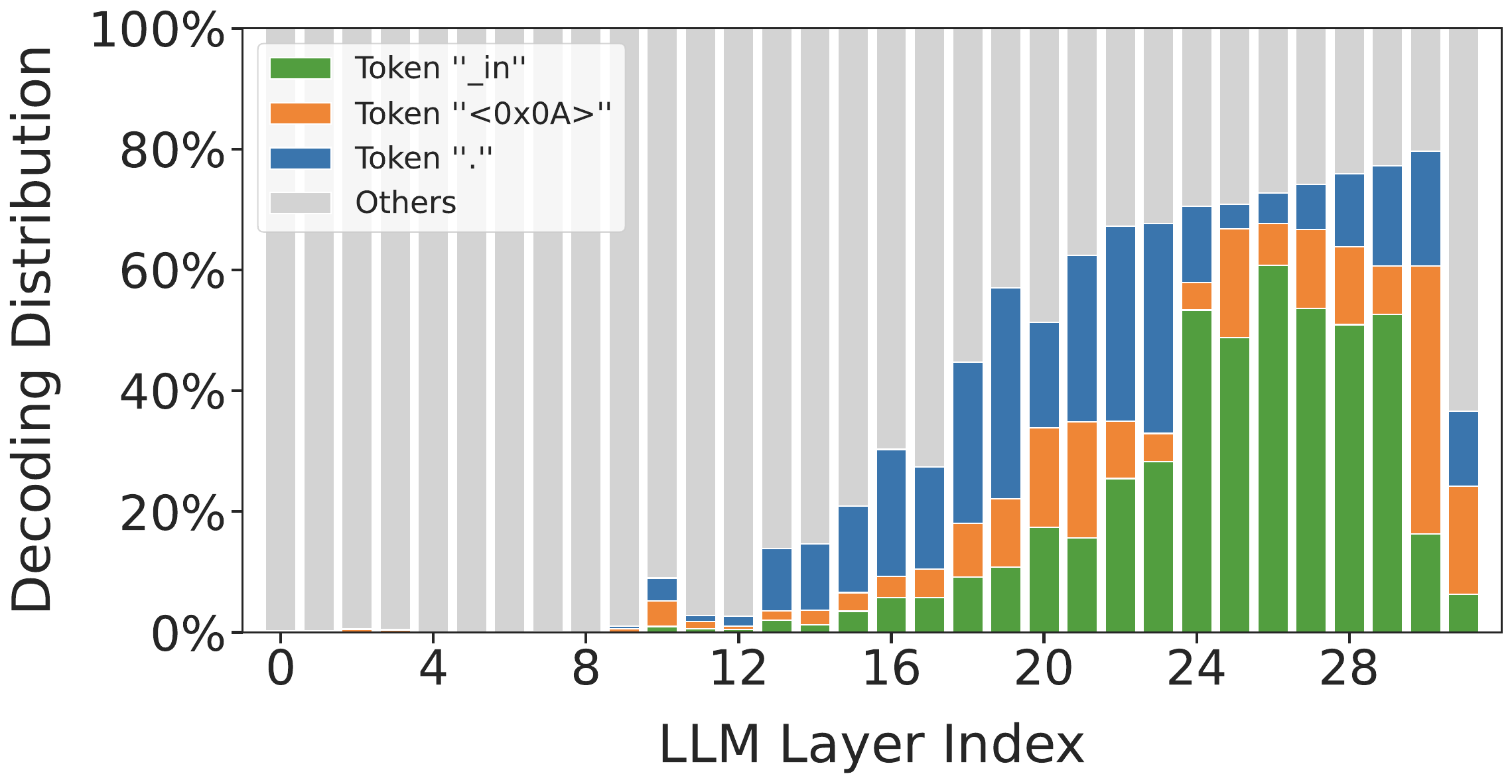}
    \vspace{-7mm}
    \caption{\textbf{Visual token decoding distributions across layers.} 
    }
    \label{fig:token_prediction_stacked_bar}
    \vspace{-6mm}
\end{figure}

\myPara{Impact of the Degraded Layer Selection.} 
In our main experiments~\cref{tab:main_results}, we apply the predictive loss $\mathcal{L}_{\text{PRe}}$(~\cref{eq:pre_loss}) to the intermediate layer of the LLM. This choice is motivated by two primary reasons. First, MLLMs tend to process and ground visual information primarily in their shallow and intermediate layers~\cite{zhang2025cross}. Second, the global semantic degradation is most pronounced in the mid-layer regions~\cref{subsec:VRD}.
To empirically validate the targeting layer, We compare with the performance of applying the PRe loss at the final layer, where the representation has been fully refined for the language task. As shown in~\cref{tab:ablation_layer_choice}, applying at the mid-layer achieves the best results. Moreover, we find that applying the PRe loss to the last layer may cause a performance drop compared to the baseline.  To investigate the cause, we perform a logit lens analysis~\cite{jiang2025devils} on the visual hidden states across different layers. As shown in~\cref{fig:token_prediction_stacked_bar}, we found that deep-layer visual representations are decoded to high-frequency, semantically ambiguous tokens such as `\_in', `.' and `$<$0x0A$>$'. We think this is a mechanism to resolve a conflict between the visual inputs and the LLM's objective. The deep layers of an LLM are aggressively optimized for next-token prediction, a task that is ill-posed for non-sequential and redundant visual patch features and could interfere with text generation. To avoid this conflict, the model learns to \textit{mute} the visual tokens in its deep layers by collapsing their representations to predict high-frequency, semantically inert tokens. Applying PRe at this stage is therefore counterproductive, as it forces the model to retain a visual structure it is actively discarding.

\begin{table}[t!]
    \centering\small
    \caption{\textbf{Results using different anchor features.}}
    \label{tab:Anchor_Source}
    \vspace{-5pt}
    \setlength{\tabcolsep}{3.5pt}
    \resizebox{\linewidth}{!}{%
    \begin{tabular}{c ccccc}
    \toprule
     Anchor Source & GQA & MMMU & VQA$^\text{T}$ & RWQA & MMVP \\
    \hline
    \rowcolor{lightgray}
    \multicolumn{6}{l}{\textit{CLIP (frozen) + Vicuna-7B-v1.5 }} \\
    Baseline & 62.0	&35.7 &45.5 & 54.8  & 20.0 \\
    Pre-LLM & 62.7& \textbf{36.1} &\textbf{46.6} & \textbf{55.4} & 22.0 \\
    Pre-Proj & 62.7 & 35.1 & 46.4 & 54.4 & \textbf{32.7}\\
    DINOv2 & \textbf{62.8} &  35.9 & 46.5 & 54.6 &28.7\\
    \hline
    \rowcolor{lightgray}
    \multicolumn{6}{l}{\textit{SigLIP2 (trainable) + Qwen2.5-7B-Instruct }} \\
    Baseline & 63.5&45.8&59.2& 60.3 & 46.0 \\
   Pre-LLM &  \textbf{64.4} &46.2& 59.7&\textbf{61.9} & 46.7 \\
    Pre-Proj & \textbf{64.4} & 45.6 & \textbf{60.1} & 60.5 & \textbf{48.7} \\
    DINOv2 & 64.1 & \textbf{46.4} & 58.9& 60.2 & 45.3\\
    \bottomrule
    \end{tabular}}
    \vspace{-3.5mm}
\end{table}

\begin{figure}[t]
    \centering
\includegraphics[width=\columnwidth]{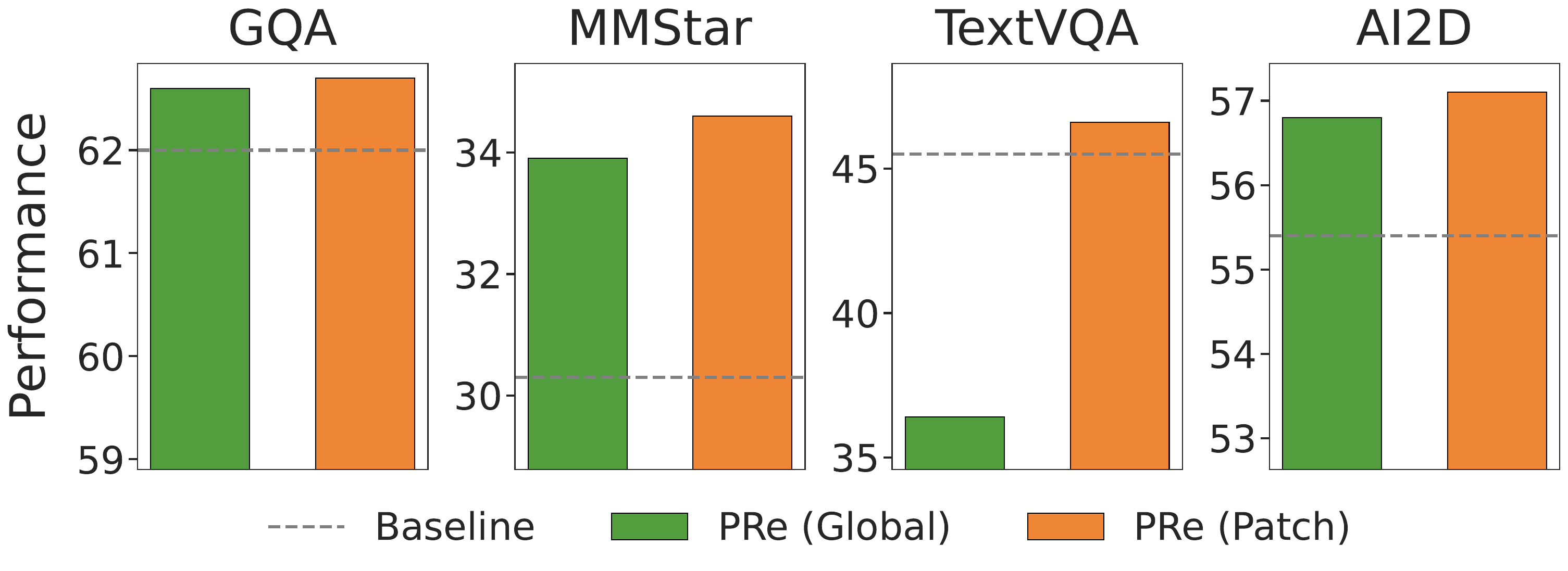}
    \vspace{-7mm}
    \caption{\textbf{Comparison of PRe applied to global features vs. patch features.} The baseline architecture employed is  CLIP-
ViT-L/14@336 (frozen) and Vicuna-7B-v1.5.}
    \label{fig:f_ablation}
    \vspace{-5.5mm}
\end{figure}

\myPara{Impact of Anchor Feature Selection.}
Another critical design choice for PRe is the selection of the anchor representation, \ie, $\mathbf{h}_{v,i}^0$ in~\cref{eq:pre_loss}, which the degraded features are tasked to predict. Our default approach uses the visual features before entering the LLM, which we denote as {Pre-LLM}. We compare this against two alternatives: (1) using the features directly from the vision encoder's output, before the projection layer, denoted as {Pre-Proj}; and (2) using features from a powerful pretrained vision foundation model, DINOv2~\cite{oquab2023dinov2}, which serves as a source of high-quality, external visual semantics.
The results are summarized in~\ref{tab:Anchor_Source}.
Firstly, using Pre-LLM features as anchors consistently yields performance improvements over the baseline. Secondly, utilizing features from Pre-Proj notably enhances performance on MMVP. This can be attributed to its direct anchoring to the visual encoder's features.
Furthermore, we find that using external DINOv2 features as the anchor, while still beneficial compared to the baseline without PRe, does not surpass our default Pre-LLM strategy. We think this is due to a representational gap between DINOv2 features space and MLLM's internal feature space. Forcing LLM to align with this externally misaligned space may introduce conflicting optimization objectives.  More importantly, current MLLMs typically perform patch merging within their projectors to alleviate computational burden, which consequently prevents the direct application of the pre-Proj method due to the difficulty in aligning feature dimensions. Using external foundation models similarly encounters this application challenge.
Collectively, this analysis of performance and practical applicability underscores the critical importance of using an internal anchor that is already a part of the model's own representational ecosystem.

\myPara{Global vs. Patch-level Regularization.}
Our PRe operates on patch-level representations, motivated by the  semantic fusion observed in our analysis. To validate this design choice, we compare it against a variant that applies the predictive loss on globally aggregated features. Specifically, instead of matching each patch feature, this variant predicts the single anchor feature vector $\bar{\mathbf{h}}_v^0$ from the single degraded feature vector $\bar{\mathbf{h}}_v^l$.
As shown in~\cref{fig:f_ablation}, the model regularized on patch-level features outperforms the one regularized on global features, showing the patch-level objective provides a much richer and more fine-grained supervisory signal that forces the model to preserve visual semantics like local details and spatial structure for every single patch.

\begin{figure}[t]
    \centering
\includegraphics[width=\columnwidth]{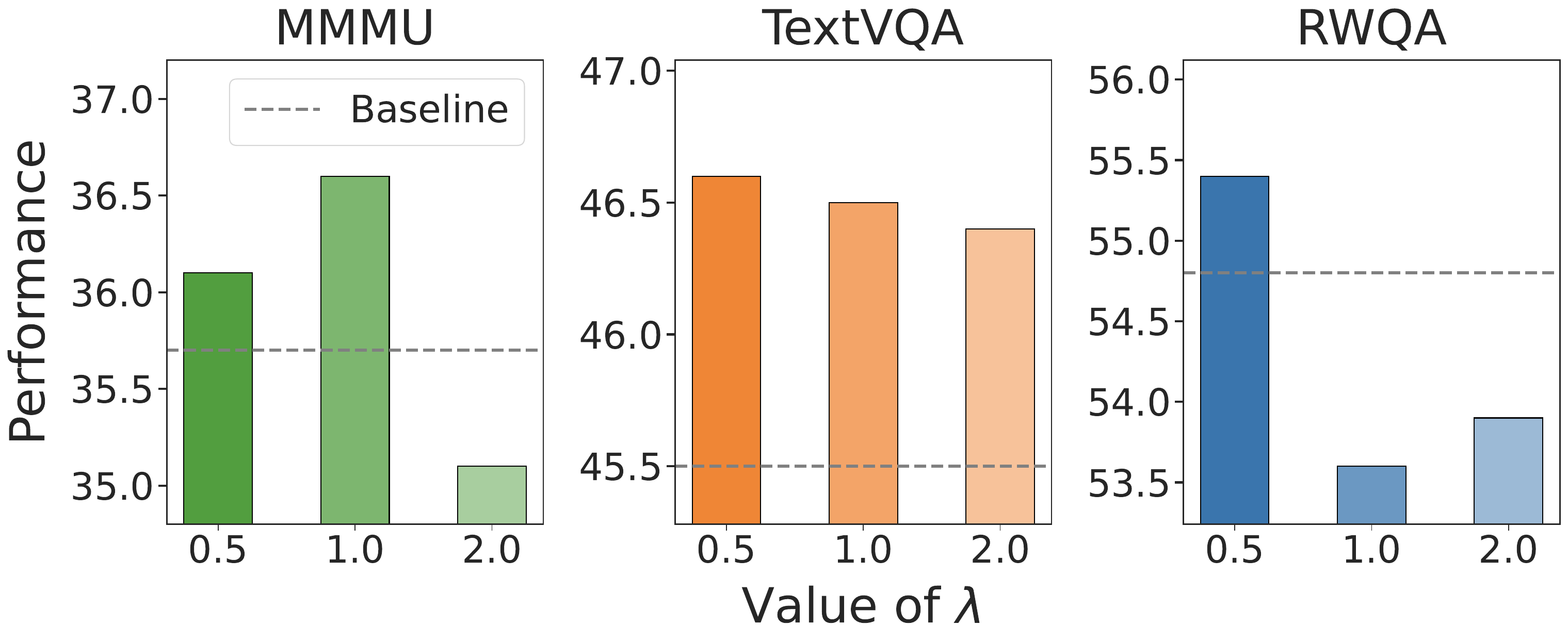}
    \vspace{-7mm}
    \caption{\textbf{The impact of $\lambda$.} The baseline architecture employed is  CLIP-ViT-L/14@336 (frozen) and Vicuna-7B-v1.5.
    }
    \label{fig:lam_ablation}
    \vspace{-5mm}
\end{figure}

\myPara{Impact of the Regularization Weight $\lambda$.}
The hyperparameter $\lambda$ in our final objective controls the strength of the predictive regularization. We conduct a sensitivity analysis of $\lambda$. As shown in~\cref{fig:lam_ablation}, 
 we evaluate the model's performance on three diverse benchmarks (MMMU, TextVQA, and RealWorldQA) with $\lambda$ set to $0.5, 1.0,$ and $2.0$.
The results show that appropriate weights can effectively improve model performance. Although increasing $\lambda$ can achieve performance improvement on some datasets, excessive regularization may interfere with the main language modeling tasks and instead reduce performance.
Based on this analysis, we set $\lambda=0.5$ as our default value for all main experiments, as it offers the best trade-off between preserving visual fidelity and maintaining language capability.

\section{Conclusion}
\label{sec:conclu}
In this paper, we first identify a critical yet underexplored issue: the degradation of visual representations in MLLMs. We observe that both global functionality and patch-level structural integrity of visual features within LLM deteriorate against their initial inputs, and attribute this to a visual sacrifice or trade-off made by the model.
We contend that this compromise leads to a brittle internal visual perception within the model.
To counteract this, we propose Predictive Regularization to maintain the visual attributes of LLM's internal representations. Experiments confirm that mitigating this degradation significantly boosts performance on various vision-language tasks, highlighting the importance of robust internal visual representations for MLLMs. We hope this research will encourage the community to establish new MLLM training strategies that concurrently optimize for intrinsic visual representation and question answering ability.
\section*{Acknowledgement}
This work is funded by  
Shenzhen Science and Technology Program (JCYJ20240813114237048),
NSFC (NO. 62206135, 62225604),
Young Elite Scientists Sponsorship Program by CAST (2023QNRC001),
the Fundamental Research Funds for the Central Universities 
(Nankai Universitiy, 070-63233085), ``Science and Technology Yongjiang 2035" key technology breakthrough plan project (2025Z053), Chinese government-guided local science and technology development fund projects (scientific and technological achievement transfer and transformation projects) (254Z0102G)


{
    \small
    \bibliographystyle{ieeenat_fullname}
    \bibliography{main}
}

\clearpage
\setcounter{page}{1}
\maketitlesupplementary

\section{Implementation Details}
\myPara{Training and evaluation configuration.}
Our training configuration simply follows llava-1.5~\cite{liu2023improvedllava} and llava-next~\cite{liu2024llavanext}. Specificall, in Stage 1, we fine-tune the projector using a learning rate of 1e-3. The total batch size is 256, achieved by distributing a per-device batch size of 16 across 8 devices with a gradient accumulation step of 2. For Stage 2, two configurations are explored: if the vision encoder is frozen, the projector and LLM are fine-tuned with a learning rate of 2e-5; alternatively, if the entire model is fine-tuned, the vision encoder's learning rate is set to 2e-6, while the remaining model components (projector and LLM) use 1e-5. Stage 2 utilizes a total batch size of 128 (per-device batch size of 4 across 8 devices, with a gradient accumulation step of 4). Common to both stages are the use of DeepSpeed-ZeRO-3 for memory optimization, a cosine learning rate scheduler, AdamW optimizer, and weight decay and warmup ratio set to 0 and 0.03, respectively. Each stage is trained for a single epoch. For PRe parameters, we set $\lambda=0.5$ as the default and select the middle layer of LLM as the target regularization layer, the ablation
experiments are discussed in the~\cref{sec:exp} of the main paper. We evaluate most datasets using lmms-eval, except for evaluating MMVP using the standard scripts provided by Cambrian-1. A simple training algorithm of our PRe is shown in~\cref{alg:pre_simple}

\begin{table*}[!t]
    \centering\small
    \caption{
    \textbf{The effectiveness of the predictive regularization on a 3B-LLM}. * indicates the visual encoder is frozen during training, without * indicates the visual encoder is trainable.
    }
    \label{tab:main_results_supp}
    \vspace{-5pt}
    \setlength{\tabcolsep}{5pt}
    \resizebox{\textwidth}{!}{%
    \begin{tabular}{lc llllllllll}
    \toprule
    \multirow{2}{*}{Visual Encoder} & \multirow{2}{*}{$\mathcal{L}_{\text{PRe}}$ }& \multicolumn{5}{c}{General \& Knowledge} & \multicolumn{2}{c}{OCR-related} & \multicolumn{2}{c}{Vision-centric} \\ 
\cmidrule(rl){3-7} 
\cmidrule(rl){8-9}
\cmidrule(rl){10-11} & & GQA & MMMU & AI2D & MMStar & SQA$^\text{I}$ & VQA$^\text{T}$ & OCRbench & RWQA & MMVP\\
\midrule
\rowcolor{lightgray}
    
    \multicolumn{11}{l}{\textit{LLM: Qwen2.5-3B-Instruct  }} \\

        \multirow{2}{*}{CLIP-ViT-L/14@336 * } & -- & 61.3 &40.7&62.2&41.3&71.9	&43.3	&318	 &52.5 &31.3 \\
    & \checkmark & \greenscore{61.3}{+0.0} &  
    \greenscore{40.9}{+0.2}&\greenscore{62.5}{+0.3} & \redscore{41.1}{-0.2}&\greenscore{72.5}{+0.6} &\greenscore{43.4}{+0.1}& \greenscore{320}{+2} &
    \greenscore{55.3}{+2.8}& \redscore{30.7}{-0.6}\\

    
\multirow{2}{*}{CLIP-ViT-L/14@336} & -- & 61.0 	&39.6 &	62.2& 42.6 &72.7	&	45.5	&359&55.6
& 38.0 \\
    & \checkmark & \redscore{60.9}{-0.1}	& \greenscore{40.0}{+0.4}&\greenscore{62.4}{+0.2}&	\greenscore{43.2}{+0.6}	&	\greenscore{72.9}{+0.2}&\greenscore{45.9}{+0.4}	&\greenscore{362}{+3}	 &\greenscore{55.8}{+0.2} & \greenscore{38.7}{+0.7}
    \\
    
    \bottomrule
    \end{tabular}}
\end{table*}

\myPara{Implementation details of analysis experiments.}
In~\cref{subsec:VRD}, we conduct extensive diagnostic experiments to point out the problem of visual representation degradation. For the \textit{Global Functional Degradation} analysis (Figure~\ref{fig:linear_probe_main}), we use the publicly available, pre-trained weights of several popular MLLMs from huggingface: Qwen-2.5-VL-7B-Instruct, LLaVA-1.5-7B, LLaVA-1.6-7B, and InternVL-3.5-4B. This allowed us to demonstrate the generality of the degradation phenomenon across different models. For the diagnostic analyses in \textit{Patch Structure Degradation} (\cref{fig:patch_contrast} and~\cref{fig:patch_similarity_heatmap}) and \textit{Degradation as a Trade-off for Language Capability} (\cref{fig:statistical_properties} and~\cref{fig:tradeoff_evolution}), we trained a baseline model employs the Qwen-2.5-7B-Instruct as LLM paired with a CLIP-ViT-L/14@336 vision encoder. The vision encoder was kept frozen during the initial pre-training stage but was made trainable during the visual instruction tuning stage. All analyses in these paragraphs were conducted on this consistently trained baseline to ensure a controlled experimental environment.

\begin{algorithm}[t]
\caption{PRe Training Step}
\label{alg:pre_simple}
\begin{algorithmic}
\State \textbf{Input:} Image batch $\mathbf{I}$, Text batch $\mathbf{T}$, MLLM $M$, PRe head $f_{pred}$.
\State \textbf{Hyperparameters:} Target layer $l_{target}$, weight $\lambda$.
\State
\textit{\textbf{1. Get initial \& intermediate visual representations}}
\State $\mathbf{H}_v^0 \gets M.\text{encode\_image}(\mathbf{I})$ \Comment{Features after projection}
\State $\mathbf{H}_t^0 \gets M.\text{tokenizer}(\mathbf{T})$ \Comment{Get text tokens}
\State $\mathbf{H}^l \gets M.\text{forward\_llm}(\mathbf{H}_v^0, \mathbf{H}_t^0 )$ \Comment{Get all layer hidden states}
\State $\mathbf{H}_v^{l_{target}} \gets \text{GetVisualTokens}(\mathbf{H}^{l_{target}})$ \Comment{Visual tokens from the target layer}
\State
\textbf{\textit{2. Define anchor (target) and online (prediction)}}
\State $\mathbf{z}_{\text{anchor}} \gets \text{stop\_gradient}(\mathbf{H}_v^0)$
\State $\mathbf{p}_{\text{online}} \gets f_{pred}(\mathbf{H}_v^{l_{target}})$
\State
\textbf{\textit{3. Compute losses}}
\State $\mathcal{L}_{\text{LM}} \gets \text{LanguageModelLoss}$
\State $\mathcal{L}_{\text{PRe}} \gets -\text{CosineSimilarity}(\mathbf{p}_{\text{online}}, \mathbf{z}_{\text{anchor}})$
\State $\mathcal{L}_{\text{total}} \gets \mathcal{L}_{\text{LM}} + \lambda \mathcal{L}_{\text{PRe}}$
\State \textbf{\textit{4. Update model parameters}}
\State $\mathcal{L}_{\text{total}}.\text{backward}()$
\State $\text{optimizer.step}()$
\end{algorithmic}
\end{algorithm}

\myPara{Details of `patch structure degradation' paragraph.}
To understand the microscopic mechanism behind the global functional degradation, we analyze the semantic structure at the patch level. Our analysis hinges on quantifying the separability of patch representations by leveraging ground-truth segmentation masks from the COCO-Stuff dataset~\cite{caesar2018cocostuffthingstuffclasses}. For each image and for the sequence of patch representations $\mathbf{H}_v^l$ at a given layer $l$, we compute the following metrics:

First, we measure the Intra-Object Cohesion, which quantifies how tightly clustered the representations of patches belonging to the same object instance are. It is defined as the average cosine similarity between all unique pairs of patches within the same object class $c$:
\begin{equation}
    \text{Cohesion}(\mathbf{H}_v^l) = \mathbb{E}_{c \in \mathcal{C}} \left[ \mathbb{E}_{i,j \in \mathcal{S}_c, i \neq j} [\cos(\mathbf{h}_{v,i}^l, \mathbf{h}_{v,j}^l)] \right]
    \label{eq:cohesion}
\end{equation}
where $\mathcal{C}$ is the set of all object classes in the image, $\cos$ is cosine similarity distance, and $\mathcal{S}_c$ is the set of patch indices belonging to class $c$. A higher cohesion value indicates that the model groups patches of the same object together more effectively.

Second, we measure the Inter-Object Coupling, which quantifies the degree of confusion or similarity between different objects. It is defined as the average cosine similarity between pairs of patches belonging to different object classes $c$ and $c'$:
\begin{equation}
    \text{Coupling}(\mathbf{H}_v^l) = \mathbb{E}_{c, c' \in \mathcal{C}, c \neq c'} \left[ \mathbb{E}_{i \in \mathcal{S}_c, k \in \mathcal{S}_{c'}} [\cos(\mathbf{h}_{v,i}^l, \mathbf{h}_{v,k}^l)] \right]
    \label{eq:coupling}
\end{equation}
A lower coupling value is desirable, as it signifies better separation between different objects.

Finally, we define the Semantic Contrast Ratio as the ratio of these two metrics. It serves as a holistic measure of patch-level separability, where a higher value is better:
\begin{equation}
    \text{Contrast}(\mathbf{H}_v^l) = \frac{\text{Cohesion}(\mathbf{H}_v^l)}{\text{Coupling}(\mathbf{H}_v^l)}
    \label{eq:contrast}
\end{equation}
These metrics allow us to quantitatively track how the semantic boundaries between and within objects evolve across the layers of the MLLM.

\myPara{Details of `degradation as a visual sacrifice for language capability' paragraph.}
In~\cref{fig:statistical_properties} (left), we quantify the {Geometric Complexity} using the PCA effective dimension. We compute the covariance matrix of the centered features and perform eigenvalue decomposition. The effective dimension is the minimum number of principal components $k$ required to explain 95\% of the total variance:
\begin{equation}
    \text{EffectiveDim}(\mathbf{X}) = \min \left\{ k \mid \frac{\sum_{i=1}^{k} \lambda_i}{\sum_{i=1}^{D} \lambda_i} \geq 0.95 \right\}
    \label{eq:pca_eff_dim}
\end{equation}
where $\lambda_i$ are the eigenvalues sorted in descending order. A higher effective dimension indicates a more complex and higher-capacity representation.

In~\cref{fig:statistical_properties} (right),  we measure the {Linear Feature Redundancy} by calculating the mean absolute off-diagonal correlation. This metric quantifies the degree of statistical dependence between feature dimensions. We compute the Pearson correlation matrix $\mathbf{C} \in \mathbb{R}^{D \times D}$ of the features, and the metric is then the average of the absolute values of its off-diagonal elements:
\begin{equation}
    \text{Redundancy}(\mathbf{X}) = \frac{1}{D(D-1)} \sum_{i \neq j} |\mathbf{C}_{ij}|
    \label{eq:redundancy}
\end{equation}
where a lower value indicates better statistical independence (\ie, less redundancy).

\section{More Experiments}

\myPara{Results on 3B LLMs.} 
In~\cref{tab:main_results} of the main paper, we demonstrated the effectiveness of our method across a wide range of 7B-LLM-based architectures.  Here, in~\cref{tab:main_results_supp}, we further supplement these findings with experimental results on 3B LLMs.  Compared to the baseline, employing PRe yields performance improvements across multiple datasets.  These results further underscore the importance of maintaining robust internal visual representations during MLLM's language-driven next-token prediction training, and prove the universally effectiveness of our PRe.

\begin{figure}[t]
    \centering
\includegraphics[width=\columnwidth]{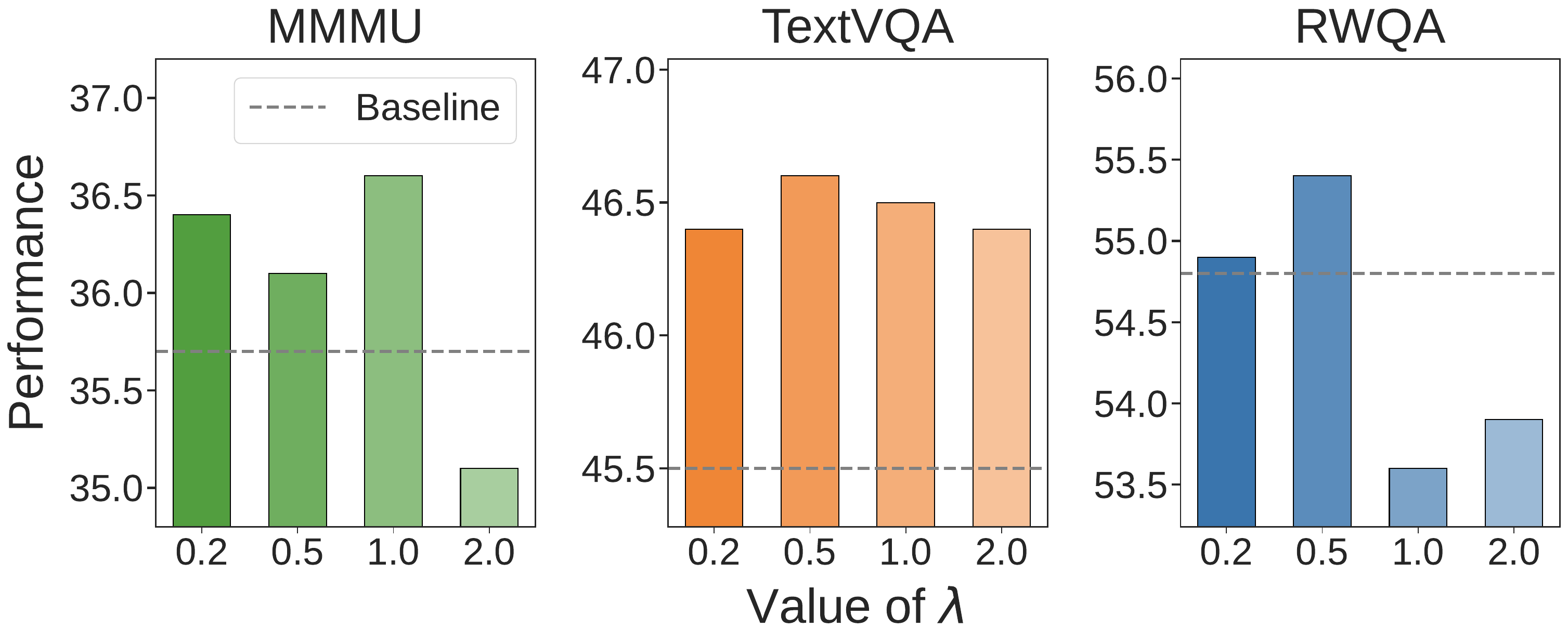}
    \vspace{-7mm}
    \caption{\textbf{The impact of $\lambda$.} The baseline architecture employed is  CLIP-ViT-L/14@336 (frozen) and Vicuna-7B-v1.5.
    }
    \label{fig:lam_ablation_supp}
\end{figure}

\myPara{Impact of the Regularization Weight $\lambda$.}
In~\cref{fig:lam_ablation} of the main paper, we show the impact of hyperparameter $\lambda$ set to $0.5, 1.0,$ and $2.0$. Here, in~\cref{fig:lam_ablation_supp}, we provide more results.  We can observe that when $\lambda$ is too small (\eg, $0.2$), the regularization is too weak to effectively counteract the representation degradation, resulting in only marginal gains over the baseline. As we increase $\lambda$, the performance improves,  however, as we further increase $\lambda$ to larger values (\eg, $2.0$), the performance begins to decline. This suggests that a moderate regularization strength is optimal, providing a sufficient signal to preserve visual fidelity without unduly interfering with the language modeling task. 
Based on this analysis, we set $\lambda=0.5$ as our default value for all main experiments, as it offers the best trade-off between preserving visual fidelity and maintaining language capability.

\begin{figure*}[t!]
    \centering
\includegraphics[width=\textwidth]{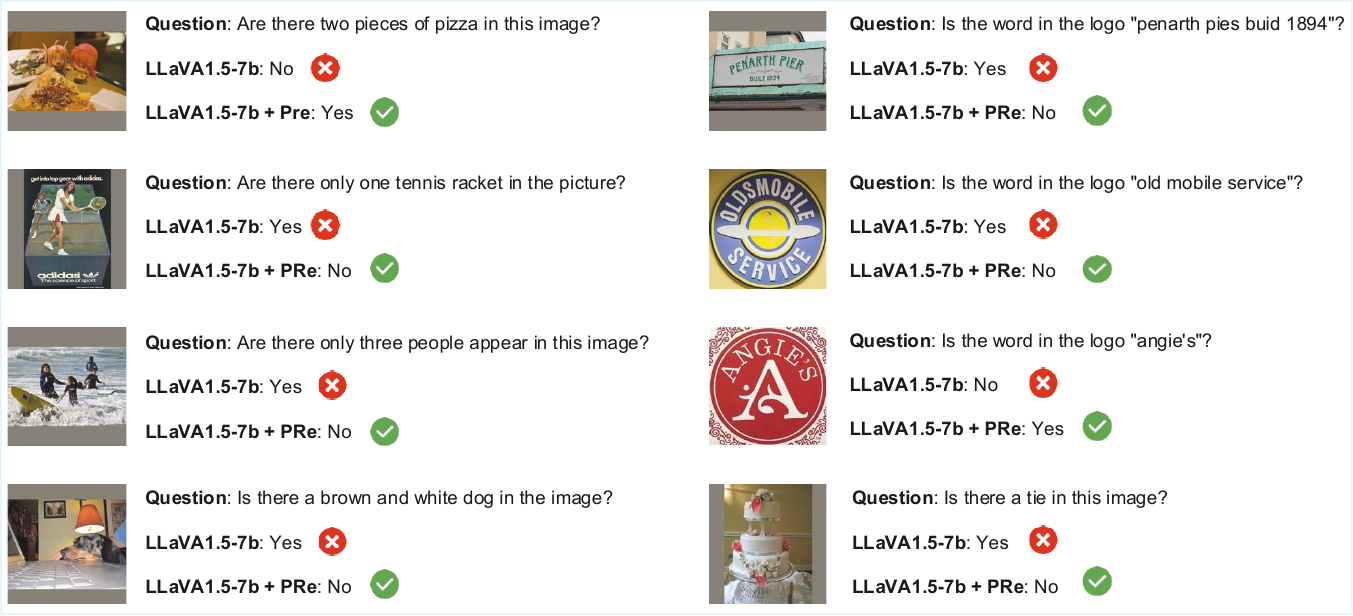}
    \vspace{-5mm}
   \caption{\textbf{Case studies comparing LLaVA-1.5-7B with and without our PRe method.}}
    \label{fig:case}
    \vspace{-3mm}
\end{figure*}

\begin{table}[t]
    \centering\small
    \caption{\textbf{Results using different anchor features.} When the CLIP is frozen, the `Pre-Proj' setting corresponds to using the penultimate layer features from a pre-trained CLIP model, together with DINOv2, DINOv3, and SAM, which form an ablation using vision foundation models}
    \label{tab:Anchor_Source_supp}
    \vspace{-5pt}
    \setlength{\tabcolsep}{3.5pt}
    \resizebox{\linewidth}{!}{%
    \begin{tabular}{l ccccc}
    \toprule
     Anchor Source & GQA & MMMU & VQA$^\text{T}$ & RWQA & MMVP \\
    \hline
    \rowcolor{lightgray}
    \multicolumn{6}{l}{\textit{CLIP (frozen) + Vicuna-7B-v1.5 }} \\
    Baseline & 62.0	&35.7 &45.5 & 54.8  & 20.0 \\
    Pre-LLM & 62.7& \textbf{36.1} &\textbf{46.6} & \textbf{55.4} & 22.0 \\
    Pre-Proj & 62.7 & 35.1 & 46.4 & 54.4 & \textbf{32.7}\\
    DINOv2-vitb14-reg4 & \textbf{62.8} &  35.9 & 46.5 & 54.6 &28.7\\
    DINOv3-vitb16  & 62.6 & 35.3 &46.1&54.9 & 28.7\\
    SAM-vitb  & 62.6 &34.2&\textbf{46.6}&54.4 & 26.0\\
    \bottomrule
    \end{tabular}}
    \vspace{-3mm}
\end{table}

\myPara{Results using different anchor features.}
In~\cref{tab:Anchor_Source} of the main paper, we present results utilizing various sources of anchor features. Here, in~\cref{tab:Anchor_Source_supp}, we further supplement these findings with results on DINOv3 and SAM. Our observations indicate that employing stronger visual foundation models, such as DINOv3, does not necessarily lead to superior performance improvements. This is likely because the feature spaces of these pure visual foundation models are optimized solely for visual tasks. Forcing the MLLM's internal multimodal feature space to predict such a pure visual space might introduce conflicting optimization objectives. Consequently, while these approaches offer a slight improvement over the baseline, their performance does not surpass our default Pre-LLM strategy. This is because Pre-LLM's features, being projections from the projector, strike a better balance between alignment with the LLM's input space and retaining robust visual capabilities.

\myPara{Case studies.} 
In~\cref{fig:case}, we present several case studies demonstrating that the application of PRe leads to notable improvements on questions related to counting, OCR, color perception, and object existence. This underscores the importance of maintaining robust internal visual features within the MLLM and validates the effectiveness of our method.

\myPara{Computational overhead.} 
As shown in~\cref{tab:com_oh}, the additional training cost is negligible, and there is zero overhead during inference since the PRe module is discarded after training. 

\begin{table}[t]
\small
\centering
\caption{\textbf{Computational overhead.}}
    \label{tab:com_oh}
    \vspace{-3mm}
    \setlength{\tabcolsep}{0.8pt}
    \resizebox{\linewidth}{!}{%
\begin{tabular}{lccc}
\toprule
\textbf{Metric} & \textbf{clip+qwen2.5-3b} & \textbf{ + PRe} & \textbf{Overhead ($\Delta$)} \\ 
\hline
Total Training PFLOPs & $2.214 $ & $2.215 $ &  $+0.045\%$\\
Throughput (samples/s) & 18.74 & 18.58 & $-0.85\%$ \\
Total Training  Time (s) & $35.5 \times 10^{3}$ & $35.8 \times 10^{3}$  &$+0.85\%$  \\
\bottomrule
\end{tabular}
}
 \vspace{-2mm}
\end{table}

\myPara{More results.} 
The main paper covers {9 diverse and general benchmarks spanning General Knowledge, OCR-related, and Vision-centric tasks}. In~\cref{tab:res_more}, we provide additional results on 5 widely-used VQA datasets. These consistent gains across total 14 benchmarks confirm the effectiveness of PRe. 

\begin{table}[t]
\small
\centering
\caption{\textbf{Results on more datasets.}}
\label{tab:res_more}
 \vspace{-3mm}
 \setlength{\tabcolsep}{2.8pt}
  \resizebox{\linewidth}{!}{
\begin{tabular}{lccccc}
\toprule
& $\text{mme}^{\text{p}}$& 	$\text{mmb}^{\text{en}}$	& $\text{mmb}^{\text{cn}}$	& $\text{seed}^{\text{i}}$	& pope(acc/f1)
\\ 
\hline
clip+qwen2.5-3b & 1408.3 &70.1 &66.1&69.9&87.5/86.3\\
\textbf{+Pre} & 1434.8 & 68.3 & 68.1 & 70.8& 88.7/87.8 \\
clip*+vicuna-7b & 1510.7 &64.3& 46.4& 58.6 &84.7/85.8\\
\textbf{+Pre} & 1528.3& 65.7 &56.2 &66.3& 86.9/85.8 \\
\bottomrule
\end{tabular}
}
 \vspace{-3mm}
\end{table}

\myPara{Generalizability to stronger encoders and higher resolutions.}
We have demonstrated the robustness of PRe across different \textbf{vision encoders} (CLIP, SigLIP2), \textbf{LLM backbones} (Vicuna, Qwen), \textbf{LLM scales} (7b, 3b), and \textbf{training paradigms }(Frozen, Unfrozen). In~\cref{tab:pre_enc}, we further validate PRe across \textbf{resolutions}, and \textbf{stronger encoders}. The consistent gains across all these settings prove the generalizability of PRe to stronger, modern baselines.
\begin{table}[t]
\small
\centering
\caption{\textbf{Results on stronger encoders and higher resolutions.} LLM: Qwen2.5-3b-Instruct.}
 \label{tab:pre_enc}
 \vspace{-2mm}
 \setlength{\tabcolsep}{2pt}
  \resizebox{\linewidth}{!}{
\begin{tabular}{lccccc}
\toprule
&GQA& AI2D	& MMStar	& $\text{VQA}^{\text{T}}$	& RWQA
\\ 
\hline
CLIP (Res: $\text{336}^{\text{2}}$) & \textbf{61.3} &62.2& \textbf{41.3}& 43.3 & 52.5\\
\textbf{+Pre} &\textbf{ 61.3}&\textbf{62.5} &41.1 &\textbf{43.4}& \textbf{55.3} \\
CLIP (Res: $\text{672}^{\text{2}}$) & 60.6&62.1& 42.6& 45.1&52.7\\
\textbf{+Pre} & \textbf{61.0}& \textbf{62.6} &\textbf{44.0} &\textbf{45.9}& \textbf{56.7} \\
Qwen2.5-VL(NaViT)  & 59.4 &60.7& 42.5& 50.0&50.5\\
\textbf{+Pre} & \textbf{60.1}& \textbf{62.2} &\textbf{43.0}&\textbf{53.2}& \textbf{51.9}\\
SigLIP2-NaFlex (NaViT)  & 60.8 &\textbf{63.1}& 43.7& 42.7&51.4\\
\textbf{+Pre} & \textbf{61.}4& 62.9 &\textbf{44.7}&\textbf{46.4}& \textbf{53.3} \\
\bottomrule
\end{tabular}
}
 \vspace{-3mm}
\end{table}

\section{Limitations and Future Work.}
In this work, we first identify the issue of visual degradation within MLLMs and subsequently propose Predictive Regularization (PRe) to maintain robust internal visual representations, thereby enhancing performance on vision-language tasks. Our approach is directly inspired by classic and influential works in visual representation learning, such as SimSiam and JEPA.
However, a potential limitation of our current PRe strategy is that using internal representations to predict the initial visual representations might inherit biases inherent to the original visual encoder. Furthermore, the broader integration of various visual representation learning methods with MLLM pre-training remains largely unexplored. For instance, the potential of contrastive learning approaches, beyond our current predictive coding-inspired method, is yet to be fully investigated within this context.
In future work, we aim to further strengthen the link between advanced visual representation learning techniques and MLLM pre-training. Our goal is to continuously optimize the visual representations within MLLMs and leverage a wider spectrum of self-supervised learning paradigms to foster even more robust and versatile multimodal models.


\end{document}